\documentclass[final]{cvpr}

\usepackage{times}
\usepackage{epsfig}
\usepackage{graphicx}
\usepackage{amsmath}
\usepackage{amssymb}
\usepackage{booktabs}
\usepackage{multirow}
\usepackage{caption} 
\usepackage[pagebackref=true,breaklinks=true,colorlinks,bookmarks=false]{hyperref}

\def\cvprPaperID{5916} 
\def\confYear{CVPR 2021}

\begin{document}
\title{Unsupervised Visual Representation Learning by Tracking Patches in Video}

\author{
Guangting Wang$^{1}$ \quad Yizhou Zhou$^{1}$ \quad Chong Luo$^{2}$ \quad Wenxuan Xie$^{2}$ \quad Wenjun Zeng$^{2}$ \quad Zhiwei Xiong$^{1}$ \\
University of Science and Technology of China$^{1}$ \qquad Microsoft Research Asia$^{2}$ \\
{\tt\small \{flylight, zyz0205\}@mail.ustc.edu.cn \quad \{cluo, wenxie, wezeng\}@microsoft.com\quad zwxiong@ustc.edu.cn}}

\maketitle
\thispagestyle{empty}

\begin{abstract}
Inspired by the fact that human eyes continue to develop tracking ability in early and middle childhood, we propose to use tracking as a proxy task for a computer vision system to learn the visual representations. Modelled on the Catch game played by the children, we design a Catch-the-Patch (CtP) game for a 3D-CNN model to learn visual representations that would help with video-related tasks. In the proposed pretraining framework, we cut an image patch from a given video and let it scale and move according to a pre-set trajectory. The proxy task is to estimate the position and size of the image patch in a sequence of video frames, given only the target bounding box in the first frame. We discover that using multiple image patches simultaneously brings clear benefits. We further increase the difficulty of the game by randomly making patches invisible. Extensive experiments on mainstream benchmarks demonstrate the superior performance of CtP against other video pretraining methods. In addition, CtP-pretrained features are less sensitive to domain gaps than those trained by a supervised action recognition task. When both trained on Kinetics-400, we are pleasantly surprised to find that CtP-pretrained representation achieves much higher action classification accuracy than its fully supervised counterpart on Something-Something dataset. Code is available online: \href{https://github.com/microsoft/CtP}{github.com/microsoft/CtP}.

    
\end{abstract}

\section{Introduction}

During the development of artificial intelligence, we can always take inspiration from the way human brain learns, and computer vision is no exception. For instance, the insight behind building the ImageNet dataset was ``to give the algorithms the kind of training data that a child was given through experiences in both quantity and quality."\footnote{Fei-fei Li's TED talk "How we teach computers to understand pictures," 2005.} In this work, we intend to address the visual representation learning problem in computer vision, so we look for clues from what developing eyes learn to do in childhood. Our intuition is that once a computer vision system learns what developing eyes are capable of, the visual features it extracts should contain the most important information needed by downstream vision tasks. 

It is not surprising that the ability to track, or to follow a moving target, caught our attention. It is not only an important capability of human eyes, but it has also been regarded as an important technology in computer vision and the basis of video analysis. In this work, however, we do not treat tracking as an ultimate task. Instead, we want to use it as a proxy task for a computer vision system to learn feature representations of visual signals. Here, the visual signals need to be videos, or moving pictures, instead of static images. Ideally, the learning process does not require human annotation, or should be self-supervised. Only in this way can we make full use of the large amount of video data on the Internet. This falls into an active area of research called self-supervised video representation learning, which aims to learn video understanding models \cite{C3D,MiCT,R3D,ProbSTFunsion} without access to human annotation.

\begin{figure}[t!]
\centering
\includegraphics[width=0.95\linewidth]{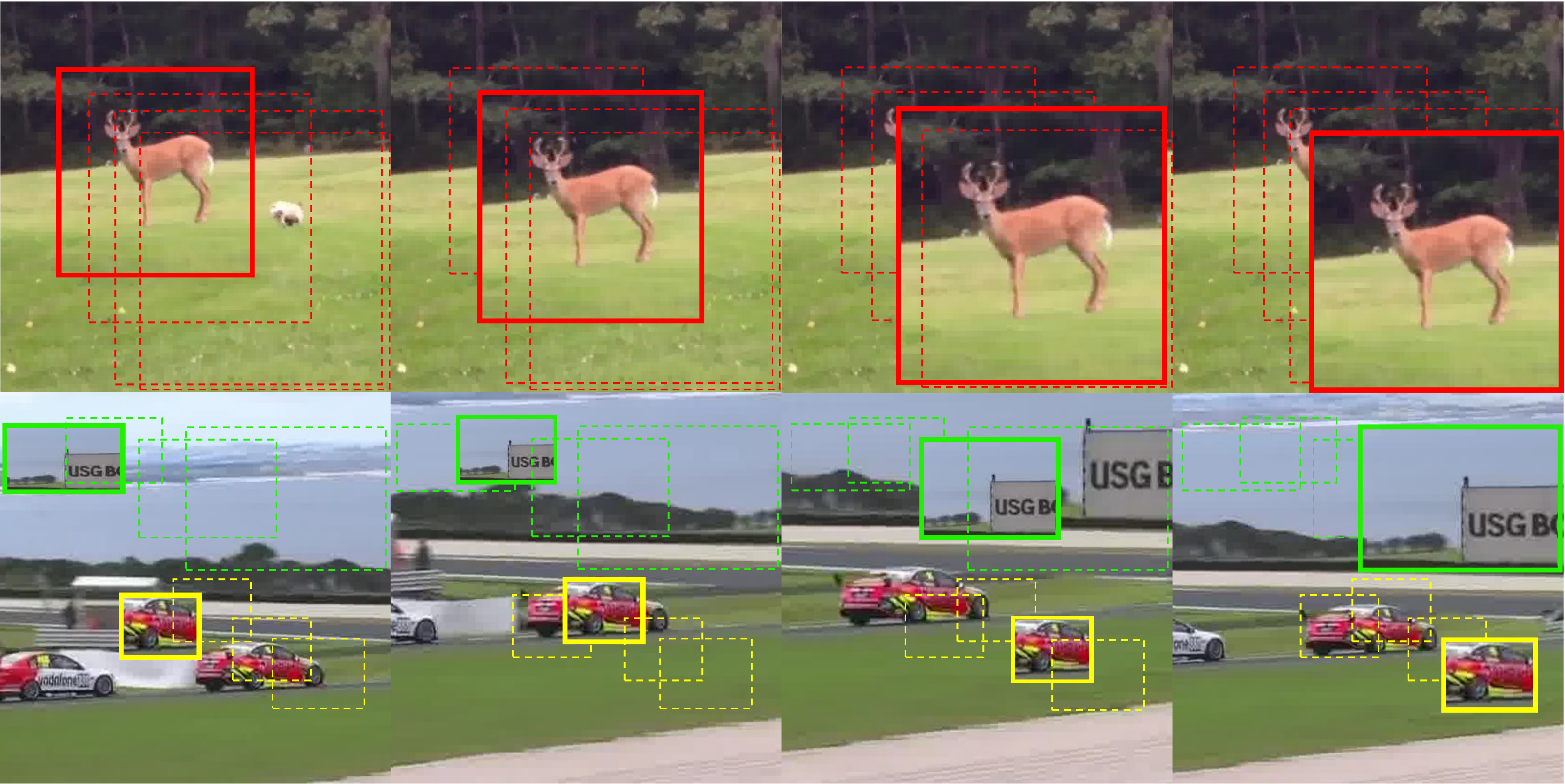}
\caption{Illustration of the Catch-the-Patch game we designed to train a computer vision system. We randomly crop one or multiple patches from a video clip, let them scale and move in a smooth way, and then train the neural network to predict the positions and sizes of the patches in each frame. }
\label{FigTemporalCorrespondence}
\end{figure}

The research progress in this area lags far behind a closely related area called self-supervised image representation learning, where several ground-breaking works \cite{MoCo, SimCLR} emerged in recent years.
A possible reason is that videos are much larger in size and more redundant in its original representation than images. It is therefore more critical to design an efficient proxy task which could guide the neural network to acquire the core capability or to distill the most important information. Some existing proxy tasks propose to estimate the orientation of video frames \cite{3DRotNet}, to predict the spatial-temporal order \cite{STPuzzle, VCOP}, or to estimate the playback speed of the input video clip \cite{SpeedNet,PRP,PlaybackSpeed}. These tasks may fail to capture the fine-grained information as they only care about the coarse global attributes. 

Our work focuses on helping the network develop the ability to follow a moving target. The proxy task used to pretrain the network is inspired by the training of human vision system. It is well-known that the \textit{Catch} game can help children develop their visual tracking abilities. For computers, we want to design a similar game. It would be ideal if we could throw all kinds of realistic objects with various appearance into the videos, but it is hard to implement. So, we step back and cut a patch from the existing video and let it change and move in the way we have pre-set it. The pretraining objective is to predict the location and size of this patch in all input frames given only the patch information in the initial frame. We call this game \textit{Catch-the-Patch} (CtP). 

Although the concept of the game is simple, it is not an easy task to design the details for the best pretraining results. We find that throwing more than one image patch, changing and moving in different patterns, in a video at the same time brings clear benefits. In addition, making image patches invisible, or disappear, from time to time can further exercise the network’s ability to associate adjacent frames. We call this masked region model (MRM). We train an R3D network \cite{R3D} and an R(2+1)D network \cite{R3D} using our invented CtP game and apply CtP-pretrained video representation to two downstream tasks, namely action recognition and video clip retrieval. 

Experimental results show that CtP significantly outperforms existing proxy tasks in video representation learning. On UCF-101 dataset, our CtP-pretrained R3D model \cite{R3D} achieves 86.2\% top-1 classification accuracy. It outperforms the most advanced method TempTrans \cite{TemporalTrans} by 6\% absolute gains. Furthermore, for datasets like Something-something-V1 which require more temporal relationship mining, CtP-pretraining leads to a 48.3\% top-1 accuracy. Surprisingly, it even surpasses the fully supervised counterpart (44.1\%) by a notable margin. To summarize, the contributions of this work are three-fold:
\begin{itemize}
	\item Inspired by the Catch game which helps children develop their eyes, we design a Catch-the-Patch game for neural networks to learn visual features from videos.
	\item We scientifically design the details of the game, including using more than one patches for training and introducing the MRM. These designs have been carefully validated by ablations studies. 
	\item We carry out comprehensive evaluation of the proposed method. CtP pretraining not only achieves state-of-the-art results for standard downstream tasks, but also closes the performance gap between unsupervised and supervised video representation learning. 
\end{itemize}

\section{Related Work}
Our work is about learning visual representations from videos, so we first review a group of most related work called unsupervised video representation learning in Section \ref{sec:related:uvrl}. Image representation learning is not involved here, as they have a very different problem setting from ours. In our proposal, tracking is used as a proxy task, but it is different from the object tracking task that computer vision researchers are familiar with. Therefore, we spend some paragraphs in Section \ref{sec:related:tracking} to discuss the connections and differences. Last, strictly speaking, the training data we used are synthetic. Can synthetic data help us achieve efficient training? We tend to have a positive answer after reviewing some related papers in Section \ref{sec:related:synthetic}. 

\subsection{Unsupervised video representation learning}\label{sec:related:uvrl}
Research works in this area fall into one of the two categories: transformation-based methods and contrastive-learning-based methods. 

The central idea of transformation-based methods is to construct some transformations so that video representation models can be trained to recognize those transformations. Typical transformations include image rotation \cite{3DRotNet}, spatial shuffling \cite{STPuzzle}, temporal shuffling \cite{VCOP}, and speed change \cite{SpeedNet,PRP,PlaybackSpeed}. Some approaches also leverage multiple transformations to improve performance. For example, VCP \cite{VCP} uses rotation degree and shuffling order as supervision signals. TempTrans \cite{TemporalTrans} integrates a set of temporal transformations including speed change and random shuffling. 

The other major category is contrastive learning \cite{DPC,MemDPC,VDIM,PacePrediction,CoCLR,VTHCL,CEP,PCL}, which has been proven effective in many other domains like image \cite{MoCo} and speech \cite{CPC} pretraining. In general, contrastive learning aims at discriminating positive and negative pairs. The definition of ``positive'' and ``negative'' pairs varies in different methods. For instance, VideoPace \cite{PacePrediction} adopts the speed attribute as the condition to assign positive and negative labels. DPC and its follow-up work \cite{DPC,MemDPC} introduces a future prediction module. The predicted future features and corresponding ground-truth future features are considered positive, while the rests are negatives. In addition to the label assigning, there is also related work that constructs training pairs between two different modalities, such as RGB-flow pair \cite{CoCLR} and audio-visual pair \cite{ACC}. 

There is no conclusion yet as to which category is better over the other, but our work falls into transformation-based methods. A major characteristic that differentiate our work from the other works in the same category is that the transformation is applied to local regions instead of the entire frame or clip. This design guides the neural network to learn region-level temporal correspondence, which we believe is the basic information for most downstream tasks. 

\begin{figure*}[t]
\centering
\includegraphics[width=0.9\linewidth]{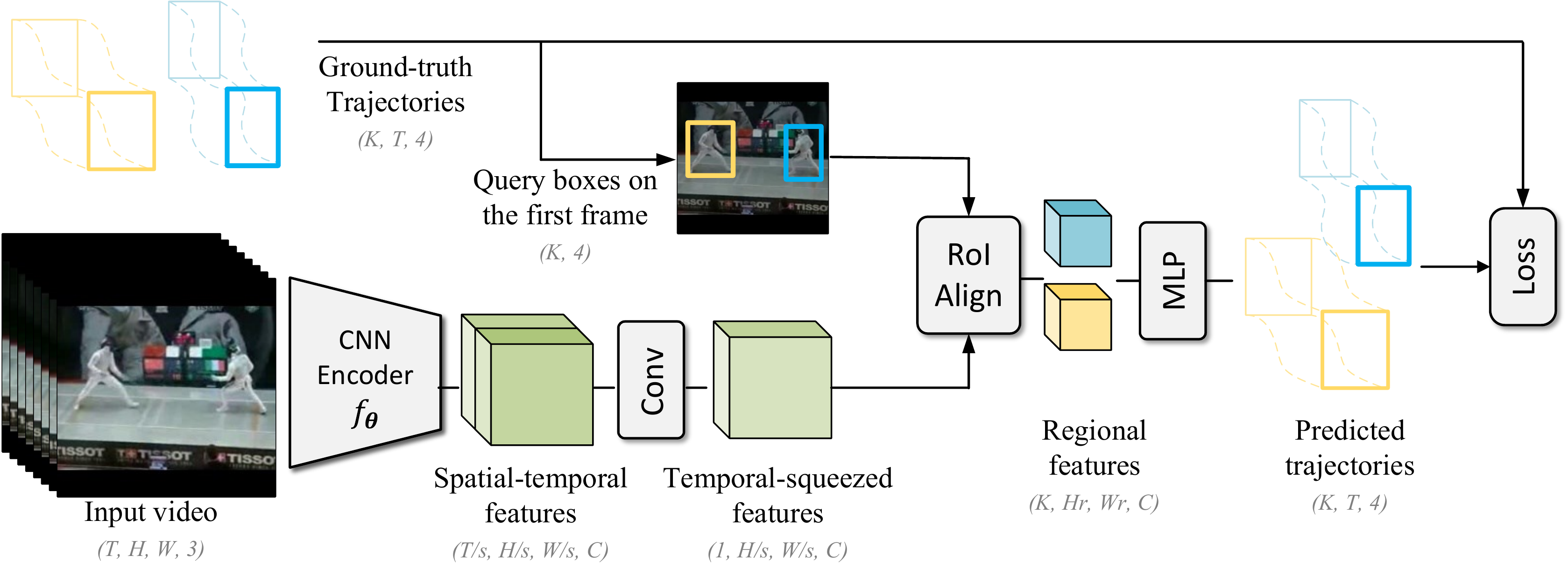}
\vspace{-6pt}
\caption{Illustration of the proposed Catch-the-Patch learning framework. The input videos are synthetic ones with overlaid image patches. The CNN encoder $f_{\boldsymbol{\theta}}$ is the 3D model we intend to pretrain. Given the initial locations of the image patches, the rest of the network is expected to predict the entire trajectories based on the extracted spatial-temporal features from $f_{\boldsymbol{\theta}}$.}
\label{FigFramework}
\end{figure*}

\subsection{Visual tracking}\label{sec:related:tracking}
Visual tracking is one of the fundamental research tasks in computer vision. There is a large body of research work that addresses both single object tracking and multiple object tracking. However, these methods are beyond the scope of this paper, as we are not trying to solve the tracking problem. Instead, we are using tracking as a proxy task to learn video representations. It should be mentioned that it is considered legitimate to leverage some non-data-driven tracking methods \cite{KCF} to provide pseudo ground truth of visual tracking in self-supervised learning. 

There are some related works \cite{TrackingColor, CycleCorrespondence, UDT, JointTaskCorr, CorrFlow, wang2015unsupervised} which use tracking as a proxy task to learn image representations. These works train a two-dimensional (2D) backbone network to extract features from a single frame. Tracking is performed between consequent frames. Since representation learning prefers an unsupervised approach to a supervised one, these works also avoid from accessing human annotations. Vondrick et al. \cite{TrackingColor} assume that the color information of a region is temporally stable, so they propose to estimate the corresponding positions based on the coherency of colors in the video. Wang et al. \cite{CycleCorrespondence} and UDT \cite{UDT} introduce a cycle-consistency constraint. After a few steps of forward tracking and then the same number of steps of backward tracking, the predicted location of the target should be close to the starting point. 

Our work is different from these image representation learning methods. Again, there is no conclusion yet whether visual representation should be learned from images or videos, but these two camps have very different problem settings and evaluation processes. The most notable difference is that video representation learning takes a video clip, or a sequence of frames, as input and trains a 3D backbone.

\subsection{Training with synthetic data}\label{sec:related:synthetic}
To meet the high demand for training data from machine learning algorithms, people produce synthetic data through various approaches, including game engines \cite{GTAV,VKITTI}, 3D models \cite{FlowNet}, and generative models\cite{LearnFromGAN}. It has been proven that synthetic data play an important role in model pretraining \cite{VKITTI,FlowNet}. The way we throw image patches into a video to create synthetic training data is similar to the idea behind Flying Chairs dataset \cite{FlowNet}. This work overlays a chair, which is generated by a 3D model with pre-set movement, on a real image. Our work overlays a cropped image patch on a real video, since 3D models are more expensive and are not able to cover all types of targets.

Our work is also related to a category of work that treats synthetic data as a regularization term. Typical work includes MixUp \cite{Mixup}, CutOut \cite{CutOut} and CutMix \cite{CutMix}. The key idea of these works is to add some constraints over the input data and labels by human priors. For example, MixUp assumes that if an image is mixed with another one, the new ground-truth label should also be a weighted combination of two original labels. Our pretraining method can also be viewed as a regularization term to the vanilla video representation model, which forces the model to encode the foreground objects' movements.


\section{Catch-the-Patch Learning Framework}

\textit{Catch}, or \textit{playing catch}, is one of the most basic children's games. The participants throw a ball, a beanbag, or a frisbee back and forth to each other. In early and middle childhood, this game helps human vision system develop the tracking ability. Now, we design the game \textit{Catch-the-Patch} for computers to develop the tracking ability. This is used as a proxy task for neural networks to learn video representations. 
In this section, we first provide a framework overview in Section \ref{SectionProblemDefinition}. Then, the design details are illustrated in Section \ref{SectionFramework}. Last but not least, we discuss how to acquire proper self-supervision signals to facilitate effective learning in Section \ref{SectionGroundTruth}.


\subsection{Framework Overview}\label{SectionProblemDefinition}


In computer vision, tracking is to locate a specified target in a video clip given the bounding box of the target in the initial frame. Target locations are usually represented by upright rectangular bounding boxes. Target locations in a sequence of frames form a tracking trajectory. In this work, we use $\boldsymbol{B}$ to denote a ground-truth trajectory:

\[\boldsymbol{B} = [\boldsymbol{b}_1, \boldsymbol{b}_2, ... ,\boldsymbol{b}_{T}],\]
\noindent where $T$ is the total number of video frames and $\boldsymbol{b}_i$ denotes the target bounding box in the $i$-th frame. 

The goal of our work is to train a general video representation model $f_{\boldsymbol{\theta}}$ parameterized by $\boldsymbol{\theta}$. The model $f_{\boldsymbol{\theta}}$ receives a video clip $\boldsymbol{\rm x}$ as input and extracts the spatial-temporal features $\boldsymbol{\rm v}$, which can be formulated as $\boldsymbol{\rm v}=f_{\boldsymbol{\theta}}(\boldsymbol{\rm x})$

In order to enable this representation to encode the information of object trajectory, we introduce a dedicated prediction head that estimates the tracking trajectory based on the extracted video representation:
\[\hat{\boldsymbol{B}} = h_{\boldsymbol{\phi}}(\boldsymbol{\rm v}, \boldsymbol{b}_1),\]
\noindent where $h$ is the prediction head and $\boldsymbol{\phi}$ is the learnable parameters. Conceptually, the function $h_{\boldsymbol{\phi}}$ takes a bounding box on the starting frame $\boldsymbol{b}_1$ as query and predicts the entire corresponding trajectory $\hat{\boldsymbol{B}}=[\hat{\boldsymbol{b}}_i]_{i=1}^{T}$. Under this formulation, we can naturally apply the ground-truth trajectory $\boldsymbol{B}$ to supervise the training of video representation model $f_{\boldsymbol{\theta}}$. Assume that we have $M$ video clips $\{\boldsymbol{\rm x}_{i}\}_{i=1}^{M}$ in the dataset and each video clip has $K$ ground-truth trajectories $\{\boldsymbol{B}_{i}^{(j)}\}_{j=1}^{K}$. The parameters $\boldsymbol{\theta}$ and $\boldsymbol{\phi}$ are jointly optimized under the loss function:
\[
  \mathcal{L} = \frac{1}{MK}\sum_{i=1}^{M}\sum_{j=1}^{K}d(\boldsymbol{B}_{i}^{(j)}, \hat{\boldsymbol{B}}_{i}^{(j)}),
\]
\noindent where $d$ is a predefined distance metric that measures how far the predicted trajectory is from the ground-truth. 

\subsection{Design details}\label{SectionFramework}

The proposed framework is composed of three components: a video representation model $f_{\boldsymbol{\theta}}$, a prediction head $h_{\boldsymbol{\phi}}$ and a distance metric $d$. In this section, we will instantiate each component. 

The detailed architecture is illustrated in Fig. \ref{FigFramework}. Generally speaking, the video representation model $f_{\boldsymbol{\theta}}$ can be a typical convolution neural network (CNN) encoder designed for video analysis tasks, such as C3D \cite{C3D}, R3D \cite{R3D} or TSM \cite{TSM}. The CNN encoder contains some temporal modules that establish relationships among video frames. The receptive field of the encoded spatial-temporal features is wide enough to cover the entire video clip, which makes it possible to explore the temporal correspondences for any queries.

By design, our prediction head $h_{\boldsymbol{\phi}}$ receives two inputs: the spatial-temporal features of the input video clip and the bounding box query on the starting frame. We adopt the RoI Align operation \cite{MaskRCNN} to associate these two inputs. RoI Align can crop the features in the given bounding boxes and encode them into a fixed size tensor. Before this operation, we use a 3D convolution layer to squeeze the temporal dimension of the spatial-temporal features. The convolution layer has a spatial kernel size of $1 \times 1$, and the temporal kernel size is the same as the temporal dimension size of the input features. This layer is motivated by the bottleneck design in ResNet \cite{ResNet}. We compress the temporal dimension and try to recover the entire trajectory from it.

After extracting regional features for each query bounding box, a two-layer multilayer perceptron (MLP) network is adopted to produce a vector of size $T \times 4$, where $T$ is the number of input frames. This vector encodes the relative deformation between the predicted trajectory and the query bounding box. Formally, we represent a bounding box $\boldsymbol{b}$ by a quadruple $(x, y, w, h)$, where $(x, y)$ is the center coordinates and $(w, h)$ is the spatial dimensions. The predicted corresponding box in the $i$-th frame $\hat{\boldsymbol{b}}_{i}$ can be written as:
\begin{alignat*}{2}
\hat{x}_i &= x_{1} + \sigma_{x} \hat{t}_{i, 1} &\quad\quad \hat{y}_i &= y_{1} + \sigma_{y} \hat{t}_{i, 2} \\
\hat{w}_i &= w_{1} \exp{(\sigma_{w} \hat{t}_{i, 2})} & \hat{h}_i &= h_{1} \exp{(\sigma_{h} \hat{t}_{i, 3})} 
\end{alignat*}

\noindent where $(x_1, y_1, w_1, h_1)$ is the query bounding box in the starting frame,  $\hat{\boldsymbol{t}}_i$ is the estimated targets for the $i$-th frame and $\boldsymbol{\sigma}$ is a set of constant scaling factors. In this work, $(\sigma_x, \sigma_h, \sigma_w, \sigma_h)$ are set to $(0.8, 0.8, 0.04, 0.04)$. 

Following the common practice in object detection \cite{FasterRCNN}, the distance function $d$ is defined in linear space for the center coordinates and log space for the spatial dimensions. Given the ground-truth bounding box $(x_i, y_i, w_i, h_i)$, we use Smooth-L1 function $L$ to calculate the distances:
\begin{alignat*}{2}
d(x_i, \hat{x}_i) &= L(\frac{x_i - \hat{x}_i}{\sigma_x}) &\quad d(y_i, \hat{y}_i) &= L(\frac{y_i - \hat{y}_i}{\sigma_y}) \\
d(w_i, \hat{w}_i) &= L(\frac{1}{\sigma_w}\log \frac{w_i}{\hat{w}_i}) & d(h_i, \hat{h}_i) &= L(\frac{1}{\sigma_h}\log \frac{h_i}{\hat{h}_i})
\end{alignat*}

Compared with the CNN encoder $f_{\boldsymbol{\theta}}$, the prediction head $h_{\boldsymbol{\phi}}$ is light-weight. For instance, an R3D-18 CNN encoder accounts for more than 82\% of the parameters in the entire framework. Hence, the power of visual tracking mainly lies in the CNN encoder. After training, the learned encoder can be applied to various downstream tasks such as video recognition, tagging, and retrieval. 

\subsection{Synthetic data generation}\label{SectionGroundTruth}

Ideally, visual representation should be learned from real objects and real trajectories. However, it is impossible to annotate the trajectories of countless objects in a huge number of videos. This is the reason why we step back and resort to synthetic data sets. 

In order to create realistic training data, we design a three-step process for data generation. First, we randomly generate a pseudo trajectory that simulates the object movement. To ensure smoothness, we first determine the bounding boxes in some key frames. The trajectory positions in the rest frames are linearly interpolated between two neighboring key frames. Second, we randomly select one bounding box from the pseudo trajectory and copy the image patch from the video frame. Finally, the copied image patch is scaled and overlaid on all original video frames according to the pseudo trajectory. When we track the copied image patch, the pseudo trajectory provides the ground-truth. We repeat this process multiple times so that each training video clip will have multiple targets and corresponding ground-truth trajectories. These targets may or may not overlap. Later, we will show through experiments that this design significantly improves the pretraining quality.


To strength the awareness of temporal relationships, we further introduce a masked region model (MRM), which is inspired by the masked language model in BERT \cite{BERT}. When constructing synthetic videos, the simulated patch will be randomly masked out in some frames with a probability of 0.2. Although the masked patch is invisible in these frames, the model is still compelled to predict the virtual locations. It encourages the pretrained model to exploit the temporal context information in successive frames.

\section{A Deep Dive into the Proxy Task}

CtP framework has adopted synthetic data for the training of proxy task. A natural concern is whether it has really learned how to track in real videos. To figure it out, we conduct an experiment to compare CtP-pretrained model with two baselines. One is a randomly initialized model, and the other is a 3D model inflated from a 2D model pretrained on ImageNet \cite{I3D}. The latter represents a model that has learned visual representations from images. We do not compare our model with standard trackers due to the big difference in problem setting. For instance, standard trackers perform tracking frame by frame, by comparing the visual features generated from 2D models. The tracking target is scaled to a fairly large resolution, and the search region is adjusted per frame according to the tracking result in the previous frame. In our case, we evaluate a 3D model by directly providing 16 sequential frames. The tracking results in these 16 frames are obtained in a single forward pass. 

\begin{figure}[t!]
\centering
\includegraphics[width=0.9\linewidth]{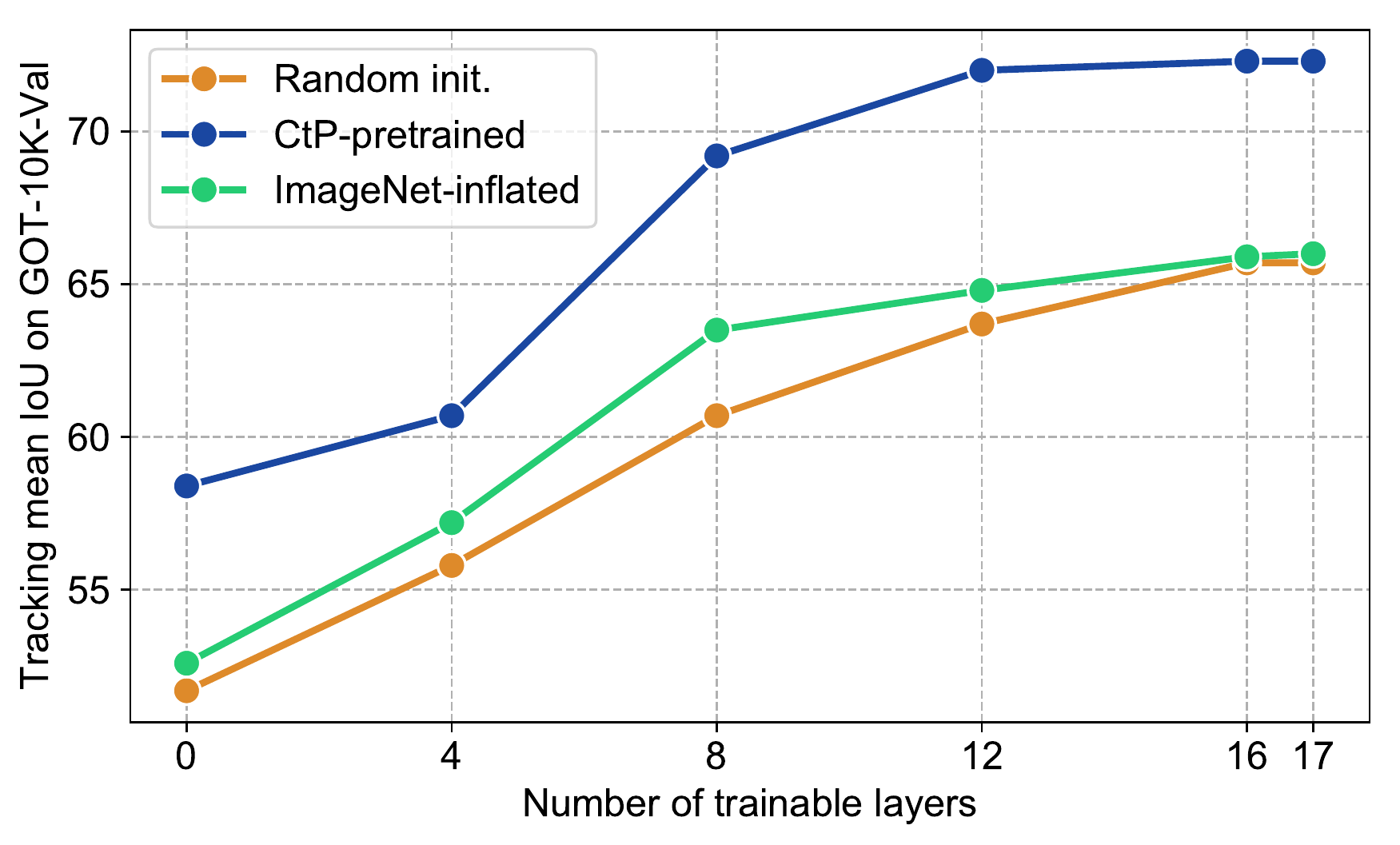}
\vspace{-4pt}
\caption{Tracking performance evaluation of a 17-layer R3D model under different training strategies.}
\label{FigTrackingIoU}
\end{figure}

We use one recently proposed large-scale tracking dataset, called GOT-10k \cite{GOT10k}, to conduct the evaluation. Specifically, we fine-tune CtP-pretrained model and two reference models using the GOT-10k training set and then evaluate them on the validation set. In the fine-tuning process, we experiment with multiple settings that freeze a different portion of parameters in the pretrained model. The evaluation metric is the mean interaction-over-union (mIoU) value between the predicted trajectory and the ground-truth. Fig. \ref{FigTrackingIoU} shows the results. CtP-pretrained model achieves significant gain over the reference models in all the fine-tuning configurations. This set of experiments verifies that our model does learn features that were beneficial to object tracking through the designed CtP game. At the very least, it provides an excellent initialization for tasks that care about object motion.

\section{Experiments}

\subsection{Implementation details}\label{SectionImpDetails}

\noindent\textbf{Model}\quad We adopt the standard R3D-18 and R(2+1)D-18 \cite{R3D} as the video representation models, following the common practice in previous research \cite{VCOP,VCP,PRP,TemporalTrans}. After the CNN encoder, the RoI Align operation produces several regional features with a spatial size of $5 \times 5$. The channel size of the MLP prediction head is 512. When transferring to the downstream tasks, we only adopt the pretrained weights of the CNN encoder and the prediction head is dropped.

\noindent\textbf{Pretraining data}\quad
Most of the evaluated models are pretrained on UCF-101 \cite{UCF101} (shorted as ``UCF'') \footnote{We use training split 1 from the official splits of UCF-101 dataset.} and Kinetics-400 datasets \cite{Kinetics} (shorted as ``K400''). During pretraining, the temporal length of input videos is 16, with a frame interval ranging from 1 to 5. For each video clip, we generate three independent ground-truth trajectories. The spatial resolution of the input clip is $112\times112$, and the size of the cropped patch is uniformly sampled from $16\times16$ to $64\times64$. The maximum speed of the trajectory is limited to 3 pixels per frame, and the scale ratio of the bounding boxes in two successive frames should fall in $[e^{-0.025}, e^{0.025}]$.

\noindent\textbf{Optimization}\quad
The pretraining process lasts for 300 epochs on UCF and 90 epochs on K400. We adopt a standard stochastic gradient descent (SGD) algorithm to optimize the training objective function. The initial learning rate is $0.01$, which is decayed by a factor of $0.1$ at 100th and 200th epoch (30th and 60th for K400), respectively. The optimizer momentum is $0.9$ and the weight decay is $10^{-4}$. During training, the total batch size is 32. Using 8 NVIDIA V100 GPU cards, training takes about 4 hours to finish on UCF and 2 days on K400.

\noindent\textbf{Evaluation}\quad We evaluate the learned video representations in two downstream tasks: action recognition and video clip retrieval. For action recognition, we append a one-layer linear classifier after the CNN encoder. The entire model is then fine-tuned on the target dataset for 150 epochs. In our experiments, we perform evaluation on UCF, HMDB-51 \cite{HMDB} (shorted as ``H51''). For video clip retrieval, we follow the same implementation as in VCOP \cite{VCOP}. In both tasks, the performance is evaluated by top-$k$ accuracy. If the class of a test video appears in its $k$ most confident predictions (or its $k$ nearest training clips for retrieval task), it is considered to be a correct classification (or retrieval). 

\subsection{Sources of ground-truths}

We have used synthetic training data in this work. The advantage is that the target trajectory is pre-set and therefore the ground-truth is always accurate and precise. The disadvantage, however, is that the pictures are not real. Apparently, there is another option to use real pictures but less precise ground-truth. In this section, we discuss and compare some alternatives to obtain pseudo ground-truth.

\textbf{Teacher tracker.}\quad One option is to generate pseudo labels by a teacher tracker. Although there have emerged many advanced trackers in the deep learning era, we choose one of the best non-data-driven trackers, named KCF \cite{KCF}, in this experiment. KCF relies on handcrafted features, so it fits into the setting of unsupervised learning.

\textbf{Cycle consistency.}\quad Cycle consistency constraint is adopted by some tracking-based unsupervised image representation learning methods \cite{CycleCorrespondence,JointTaskCorr}. We can also make use of this constraint to generate ground-truths. A simple implementation is to use the inverse trajectory of the backward tracking as the ground-truth of forward tracking. However, using this strategy alone may lead to a trivial solution of static trajectory. Therefore, we use a mixed solution where a quarter of the training labels are sourced from the synthetic data, and the rest is from backward tracking.

\textbf{Real annotations.}\quad There exist some annotated datasets for visual object tracking. We find that GOT-10k dataset \cite{GOT10k} is fairly large with a variety of object classes. The ground-truth labels are considered accurate, but one disadvantage is that there is only one annotated object in each video. Besides, the movement of non-object is not annotated throughout the dataset. 

We train R3D-18 models using these different sources of ground-truths. The learned video representations are transferred to the action recognition task on UCF and H51 datasets, as specified in the evaluation protocol \ref{SectionImpDetails}. 

Table \ref{TableSourceGT} presents the results. It is not surprising that all pretrained models achieve better performance than a randomly initialized model. When pretrained on UCF, using pre-set trajectories achieves significantly better results than the other two choices. It suggests that accurate ground-truths might be more crucial than realistic pictures for the tracking proxy task.  


We also compare between using synthetic data with pre-set trajectories and real data with human annotations. Since there is only one annotated object in each video, we also tested a setting that only uses one image patch per video, which is denoted by ``Pre-set $\times 1$''. We are surprised to find that even with this setting, there is no disadvantage in using synthetic data. The good performance may attribute to the fact that both the image patches and their trajectories can change in every training epoch. Furthermore, if we use more pre-set trajectories per video (3 as default), the performance is even better. 



\begin{table}[!t]
\caption{Ablation analysis of different ways to acquire ground-truths. In this experiment, we adopt R3D-18 models transferred to the action recognition task.} \label{TableSourceGT}
\centering
\begin{tabular}{@{}c|c|cc@{}}
\toprule
Pretraining & Source of         & \multicolumn{2}{c}{Top-1 Acc. (\%)} \\ \cmidrule(l){3-4} 
dataset     & ground-truths     & UCF          & H51          \\ \midrule
None        & -                 & 65.0             & 30.9             \\ \midrule
UCF-101     & Teacher tracker   & 70.6             & 44.4             \\
UCF-101     & Cycle consistency & 81.1             &    53.3            \\
UCF-101     & Pre-set ($\times 3$)   & 83.9             & 53.6             \\ \midrule
GOT-10k     & Real annotation  & 77.0             & 51.4              \\
GOT-10k     & Pre-set ($\times 1$) & 80.4     & 49.4              \\
GOT-10k     & Pre-set ($\times 3$)& \textbf{85.9}        & \textbf{55.7}                 \\ \bottomrule
\end{tabular}
\end{table}

\subsection{Ablation analysis}

We analyze several design choices in the Catch-the-Patch framework. For time efficiency, models evaluated in this subsection are trained on a subset (about 25\%) of K400. The pretrained models are transferred to the action recognition task on UCF \cite{UCF101} and H51 \cite{HMDB}. 

\textbf{Number of pre-set trajectories.}\quad
For each video clip, we generate different numbers of pre-set trajectories. The ablation results are presented in Table \ref{TableAblation} (a). As the number of trajectories increases from 1 to 2, the top-1 accuracy of the transferred model is significantly improved. We believe that, under single-trajectory supervision, the model tends to learn only a global motion. Using multiple trajectories simulates a situation where each region can have its own motion state. The extracted features should contain enough information for the tracking head to simultaneously capture the motion and distinguish between different trajectories. When the number of trajectories exceeds 3, the performance starts to saturate. Therefore, we use 3 trajectories in the rest of our experiments.


\textbf{Masked region model.}\quad
When generating synthetic videos, some overlaid patches are masked out with a random probability. To predict the virtual positions of masked patches, the model needs to exploit the temporal context information. In Table \ref{TableAblation} (b), we present the experimental results of training with and without MRM. It clearly shows that MRM helps to improve the video representation learning in both R3D and R(2+1)D backbones.

\begin{table}[!t]
    \caption{Ablation analysis of our proposed CtP. The model is pretrained on a subset of K400 and transferred to the action recognition task.} \label{TableAblation}
    \begin{minipage}{.4\linewidth}
      \centering
      \small
      \setlength\tabcolsep{3pt}
      \caption*{(a) Num of trajectories.}
      \begin{tabular}{c|cc}
      \toprule
      Num of     & \multicolumn{2}{c}{Top-1 Acc. (\%)} \\
       trajs & UCF          & H51          \\ \midrule
   1            & 81.7             & 49.1             \\
   2            & 83.0             & 51.9             \\
   3            & 82.8             & 53.2             \\  
   4            & 82.2             & 54.1  \\ \bottomrule 
      \end{tabular}
    
    \end{minipage}
    \begin{minipage}{.6\linewidth}
      \centering
      \small
      \setlength\tabcolsep{3pt}
      \caption*{(b) Masked region model.}
      \begin{tabular}{c|c|cc}
\toprule
\multirow{2}{*}{Backbone} & \multirow{2}{*}{MRM} & \multicolumn{2}{c}{Top-1 Acc. (\%)} \\
        &  & UCF & H51 \\ \midrule
R3D     &  & 82.8    & 53.2    \\
R3D     & \checkmark & \textbf{84.0}    & \textbf{55.3}    \\ \hline
R(2+1)D &  & 85.1    & 55.9    \\
R(2+1)D & \checkmark & \textbf{87.2}    & \textbf{57.8}    \\ \bottomrule
\end{tabular}
    \end{minipage}
\end{table}
\begin{table}[t]
\caption{Comparison with baseline pretraining approaches. We report the top-1 accuracy of transferred video representation models on UCF-101, HMDB-51 and Something-Something-V1 (SS) datasets.}
\label{TableBaseline}
\centering
\small
\setlength\tabcolsep{5pt}
\begin{tabular}{c|cc|ccc}
\toprule
\multirow{2}{*}{Backbone} & \multicolumn{2}{c|}{Pretraining} & \multicolumn{3}{c}{Top-1 Acc. (\%)} \\
        & Method                          & Dataset  & UCF           & H51 & SS            \\ \midrule
R3D     & \multicolumn{1}{c|}{None}       & None     & 65.0          & 30.9 & 39.2          \\
R3D     & \multicolumn{1}{c|}{Supervised} & ImageNet & 79.5          & 40.0 & 42.9          \\
R3D     & \multicolumn{1}{c|}{Supervised} & K400     & \textbf{91.6} &  \textbf{60.5}    & 43.3          \\
R3D     & \multicolumn{1}{c|}{CtP}        & K400     & 86.2          & 57.0 & \textbf{44.2} \\ \midrule
R(2+1)D & \multicolumn{1}{c|}{None}       & None     & 67.0          & 29.5 & 40.6          \\
R(2+1)D & \multicolumn{1}{c|}{Supervised} & K400     & \textbf{92.7} & \textbf{64.5}     & 43.9          \\
R(2+1)D & \multicolumn{1}{c|}{CtP}        & K400     & 88.4          & 61.7 & \textbf{48.3} \\ \bottomrule
\end{tabular}
\end{table}

\begin{table}[]
\caption{Comparison with state-of-the-art video representation learning approaches. The downstream task is action recognition on UCF-101 and HMDB-51 datasets. The column ``Arch.'' denote the input spatial resolution and the encoder architecture. The mark \dag \ means that the results are produced by our re-implementation.}
\label{TableSOTA}
\centering
\begin{tabular}{l|c|c|cc}
\toprule
\multirow{2}{*}{Method} & \multirow{2}{*}{Dataset} & \multirow{2}{*}{Arch.} & \multicolumn{2}{c}{Top-1 Acc. (\%)} \\ 
          &      &         & UCF  & H51 \\ \midrule
DPC \cite{DPC}      & K400  & R-2D3D  & 75.7 & 35.7 \\ 
CBT \cite{CBT}      & K600  & S3D     & 79.5 & 44.6 \\ 
MemDPC \cite{MemDPC}  & K400  & R-2D3D  & 78.1 & 41.2 \\ 
SpeedNet \cite{SpeedNet} & K400  & S3D-G   & 81.1 & 48.8 \\ 
CEP \cite{CEP} & K400 & SlowFast & 77.0 & 36.8 \\
CoCLR \cite{CoCLR} & K400  & S3D   & 87.9 & 54.6 \\ \midrule
VCP \cite{VCP}      & UCF   & R3D     & 66.0 & 31.5 \\ 
PRP \cite{PRP}      & UCF   & R3D     & 66.5 & 29.7 \\ 
TempTrans \cite{TemporalTrans}& UCF  & R3D     & 77.3 & 47.5 \\ 
Ours      & UCF   & R3D     & \textbf{83.9} & \textbf{53.6} \\ \midrule
TempTrans \cite{TemporalTrans} & K400  & R3D     & 79.3 & 49.8 \\ 
MoCo \dag\cite{MoCo}     & K400  & R3D     & 77.0 & 43.4     \\ 
VCOP \dag\cite{VCOP}      & K400  & R3D     & 73.3 & 41.4      \\ 
3DRotNet \dag\cite{3DRotNet} & K400  & R3D     & 77.5 & 41.4     \\ 
MemDPC \dag\cite{MemDPC}   & K400  & R3D     & 75.3 & 41.2     \\ 
SpeedNet \dag\cite{SpeedNet} & K400  & R3D     & 83.5 & 50.6     \\ 
Ours      & K400  & R3D     & \textbf{86.2} & \textbf{57.0} \\ \midrule
Pace \cite{PacePrediction} & UCF & R(2+1)D & 75.9 & 35.9 \\
VCOP \cite{VCOP}      & UCF   & R(2+1)D & 72.4 & 30.9 \\ 
VCP \cite{VCP}      & UCF   & R(2+1)D & 66.3 & 32.2 \\ 
PRP \cite{PRP}      & UCF   & R(2+1)D & 72.1 & 35.0 \\ 
TempTrans \cite{TemporalTrans} & UCF   & R(2+1)D & 81.6 & 46.4 \\ 
Ours      & UCF  & R(2+1)D & \textbf{86.2} & \textbf{57.1} \\ \midrule
Pace \cite{PacePrediction} & K400 & R(2+1)D & 77.1 & 36.6 \\
Ours      & K400 & R(2+1)D & \textbf{88.4} & \textbf{61.7} \\ \bottomrule
\end{tabular}
\end{table}

\subsection{Comparison with baseline approaches}

We compare our pretraining method with two baseline approaches: random initialization and supervised pretraining on Kinetics or ImageNet. After supervised pretraining on ImageNet, we inflate the 2D convolutional kernels to 3D as in I3D \cite{I3D}. The results of the action recognition task are presented in Table \ref{TableBaseline}.

CtP Pretraining achieves much higher accuracy than random initialization and ImageNet pretraining on all three datasets. On UCF-101, CtP-pretrained R3D-18 model gets an absolute gain of 20.4\% over the random baseline and 5.9 \% over the ImageNet baseline. Unsurprisingly, the model pretrained with action recognition labels on K400 achieves the highest accuracy on UCF and H51. It is encouraging that the performance gap between our model and this K400-supervised model is not large. Interestingly, when both models are evaluated on the Something-Something \cite{SomethingSomething} dataset, our model achieves a better performance. This may due to the fact that accurate classification of the fine-grained actions in SS relies on the quality of local features, which is the advantage of our method. 

\subsection{Comparison with state-of-the-art approaches}

Following common practices, we compare our method with the state-of-the-art (SOTA) approaches by transferring the learned representations to two downstream tasks. 

\begin{table}[]
\caption{Comparison with state-of-the-art video representation learning approaches in video clip retrieval task. In this experiment, we use R3D-18 as the CNN encoder.}
\label{TableRetrieval}
\small
\setlength\tabcolsep{5pt}
\centering
\begin{tabular}{l|c|cc|cc}
\toprule
\multirow{2}{*}{Method} & \multirow{2}{*}{Dataset} & \multicolumn{2}{c|}{UCF Acc. (\%)} & \multicolumn{2}{c}{H51 Acc. (\%)} \\ 
                        &                          & Top-1            & Top-5           & Top-1            & Top-5            \\ \midrule
VCOP \cite{VCOP}                   & UCF                      & 14.1             & 30.3            & 7.6              & 22.9             \\
VCP \cite{VCP}                    & UCF                      & 18.6             & 33.6            & 7.6              & 24.4             \\
PRP \cite{PRP}                    & UCF                      & 22.8             & 38.5            & 8.2              & 25.8             \\
Pace \cite{PacePrediction}                    & UCF                      & 19.9             & 36.2            & 8.2              & 24.2             \\
Ours                    & UCF                      & \textbf{23.4}             & \textbf{40.9}                & \textbf{11.4}             & \textbf{30.2}                 \\ \midrule
SpeedNet \cite{SpeedNet}                & K400                     & 13.0             & 28.1            & -                & -                \\
TempTrans \cite{TemporalTrans}             & K400                     & 26.1             & \textbf{48.5}            & -                & -                \\
Ours                    & K400                     & \textbf{29.0}             &  47.3            & \textbf{11.8}             & \textbf{30.1}                 \\ \bottomrule
\end{tabular}
\end{table}

\textbf{Action recognition.}\quad
The evaluation results are presented in Table \ref{TableSOTA}. It should be noted that the finetuning settings, including input resolution, training epochs, and data augmentations, can dramatically affect the final accuracy. Unfortunately, there is no standard setting exists. In order to present an apple-to-apple comparison, we have tried our best to integrate the existing open-sourced work with our finetuning pipeline (marked as \dag\ in Table \ref{TableSOTA}).

Overall, the CtP learning framework significantly outperforms existing approaches under the same training configurations. For example, when an R3D-18 encoder is pretrained on K400 dataset, CtP improves the very recent approach SpeedNet \cite{SpeedNet} by an absolute gain of 2.7 \% on UCF-101. Meanwhile, we also benchmark MoCo \cite{MoCo}, a representative method designed for the image representation learning, on the action recognition task. Experimental results demonstrate that it cannot work well for video. Compared with other methods trained with different architectures or resolutions, our method achieves a vastly higher accuracy of 88.4 \% on UCF-101 and 61.7 \% on HMDB-51.

\textbf{Video clip retrieval.}\quad We use the exact same evaluation protocol as in VCOP \cite{VCOP} and report the retrieval accuracy on UCF-101 and HMDB-51 datasets. The results in Table \ref{TableRetrieval} clearly shows that our pretraining method achieves superior performances on both datasets.

\subsection{Data efficiency}

\begin{figure}[t!]
\centering
\includegraphics[width=0.9\linewidth]{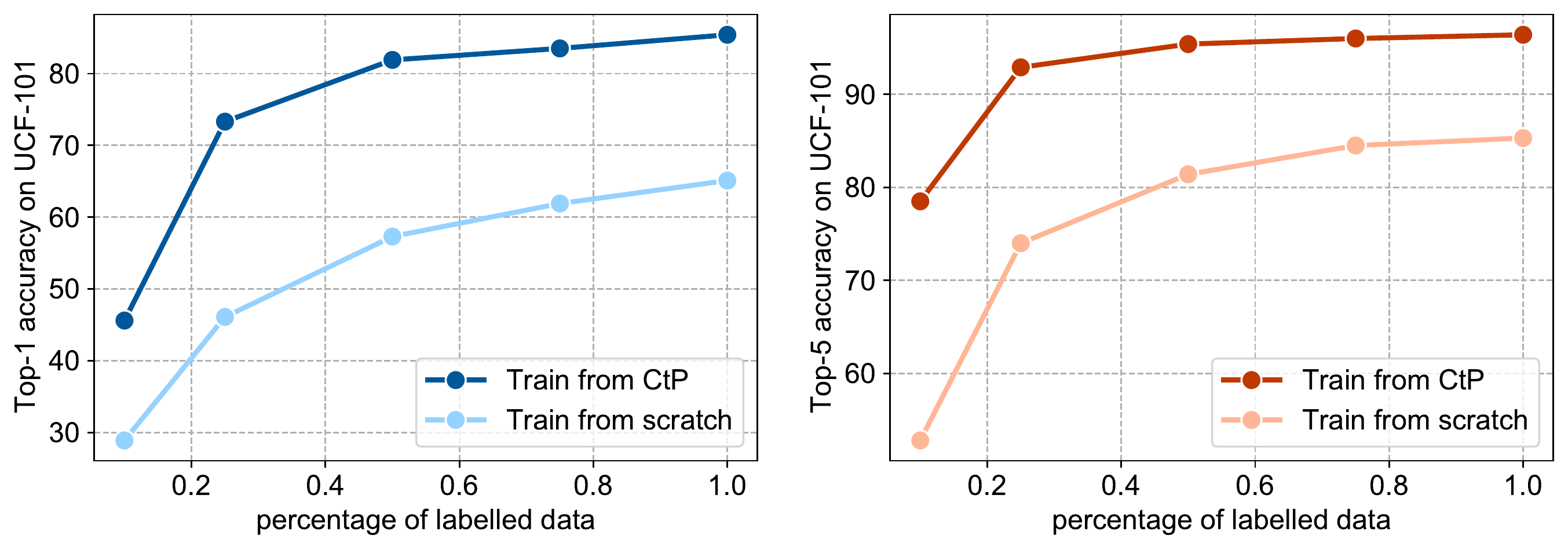}
\vspace{-8pt}
\caption{Data efficiency of representations. The pretraining is conducted on K400 dataset.}
\label{FigDataEfficiency}
\end{figure}

We plot the data efficiency curve in Fig. \ref{FigDataEfficiency}. Two models are trained with different percentages of labeled data on UCF-101. One is initialized by CtP-pretrained representations while the other is trained from scratch. The advantage of CtP-pretraining is more significant when there are fewer number of labeled data. Notably, under the help of pretraining, with only 20\% of the labeled data, we can achieve a similar performance as a randomly initialized classifier trained on the entire labeled dataset.

\vspace{-8pt}
\begin{figure}[t]
\centering
\includegraphics[width=0.9\linewidth]{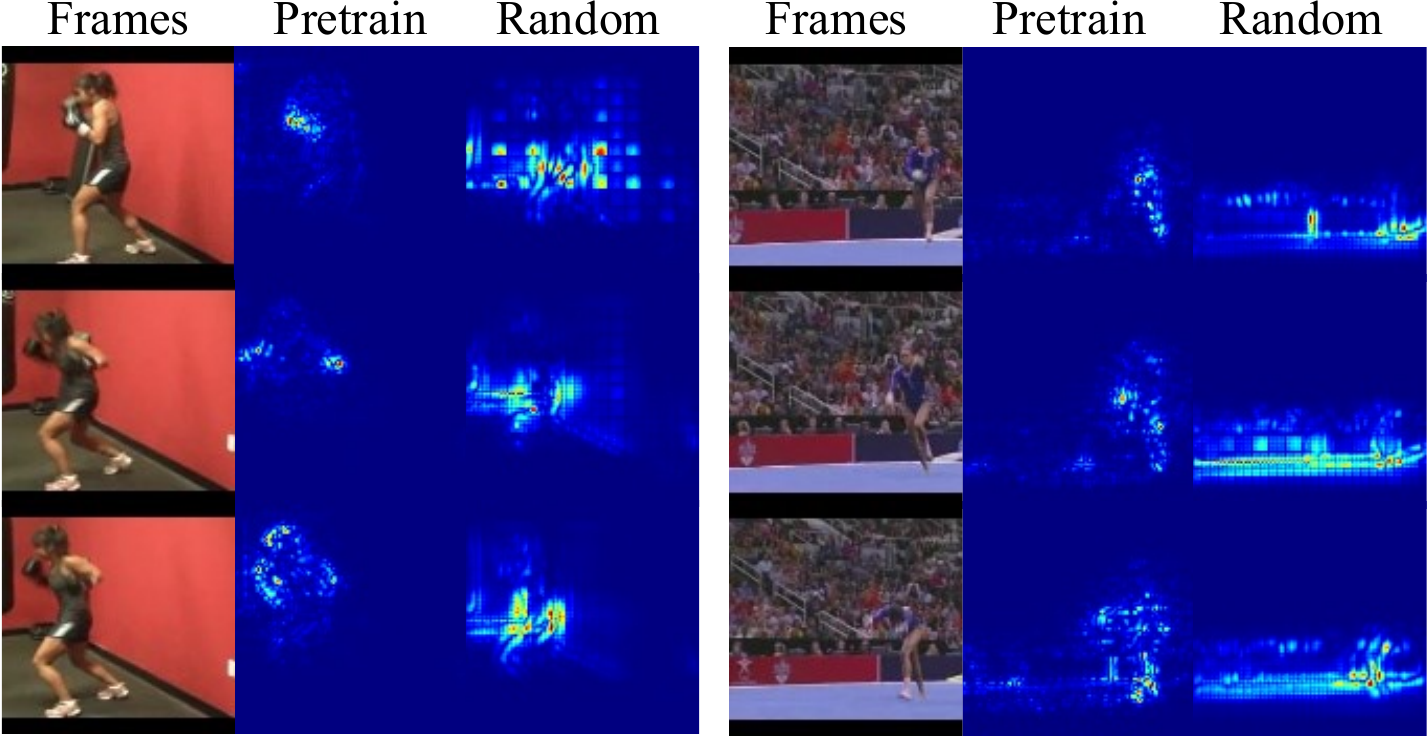}
\vspace{-8pt}
\caption{Visualization of important pixels by GradCAM}
\label{FigGradcam}
\end{figure}

\subsection{Visualization}

We use guided GradCAM \cite{GradCAM} to highlight the important pixels that contribute to the final classification decision. The classifier is pretrained and then finetuned on UCF-101. We also visualize the results of the classifier trained from scratch. Fig. \ref{FigGradcam} shows that pretrained classifier successfully captures the salient regions, while the results of the random baseline are noisy, especially for complex scenes. This further verifies that our pretraining enhances the network's ability to follow moving targets.

\section{Conclusion}

In this paper, we have proposed \textit{Catch-the-Patch} learning framework which uses tracking as a proxy task to learn video feature extraction. It is an unsupervised framework without access to any human annotations. Comprehensive experiments have proved the rationality and effectiveness of our approach. In the future, we plan to explore semi-supervised approaches and design a bootstrap process to create more realistic training data. We also plan to train the model using the sheer amount of video data on the Internet. After all, that is what unsupervised learning is all about.

\section*{Acknowledgement}

The authors would like to express their appreciation to Xiaoyi Zhang for the inspiring discussions.

\appendix

\section*{Appendix}

\section{Evaluation Pipeline}

In this section, we introduce the evaluation details when transferring the pretrained models to the downstream tasks. 

\noindent\textbf{Action recognition}\quad We follow the common practice \cite{MMAction} to construct an action recognition model. Specifically, we adopt a CNN encoder to extract the spatial-temporal feature from the raw video clip. The CNN encoder can be initialized from the pretrained models of various proxy tasks, e.g., VCOP \cite{VCOP}, 3DRot \cite{3DRotNet} or our CtP. The feature extracted by the CNN encoder is averaged in both spatial and temporal dimensions, which leads to a feature vector of size 512 (for R3D-18 and R(2+1)D-18 in our work). We append a one-layer linear classifier based on the averaged feature.

The entire action recognition model is finetuned on the target datasets. The input video clip has a temporal coverage of 32 frames. For UCF-101 and HMDB-51 datasets, we sample 16 frames with a temporal stride of 2. For Something-Something dataset, we successively sample 32 frames since this dataset emphasizes more fine-grained temporal relationships. The standard data augmentation strategy is applied in the training period, including random cropping, horizontal flip and color jitters. During inference time, the video frames are resized to a spatial resolution of $171 \times 128$, and we center crop the regions of shape $112 \times 112$. For each video, we temporal-uniformly pick 10 clips. The final classification decision of the video is averaged by the prediction results of these clips. 

The optimization lasts for 150 epochs with standard SGD algorithms. The learning rate is started from 0.01 and it is gradually decayed by a factor of 0.1 at 60th epoch and 120th epoch. The weight decay and the momentum value is set to $5\times 10^{-4}$ and 0.9, respectively. After the optimization, we adopt the model produced in the last epoch for evaluation.

\begin{figure}[t]
\centering
\includegraphics[width=0.9\linewidth]{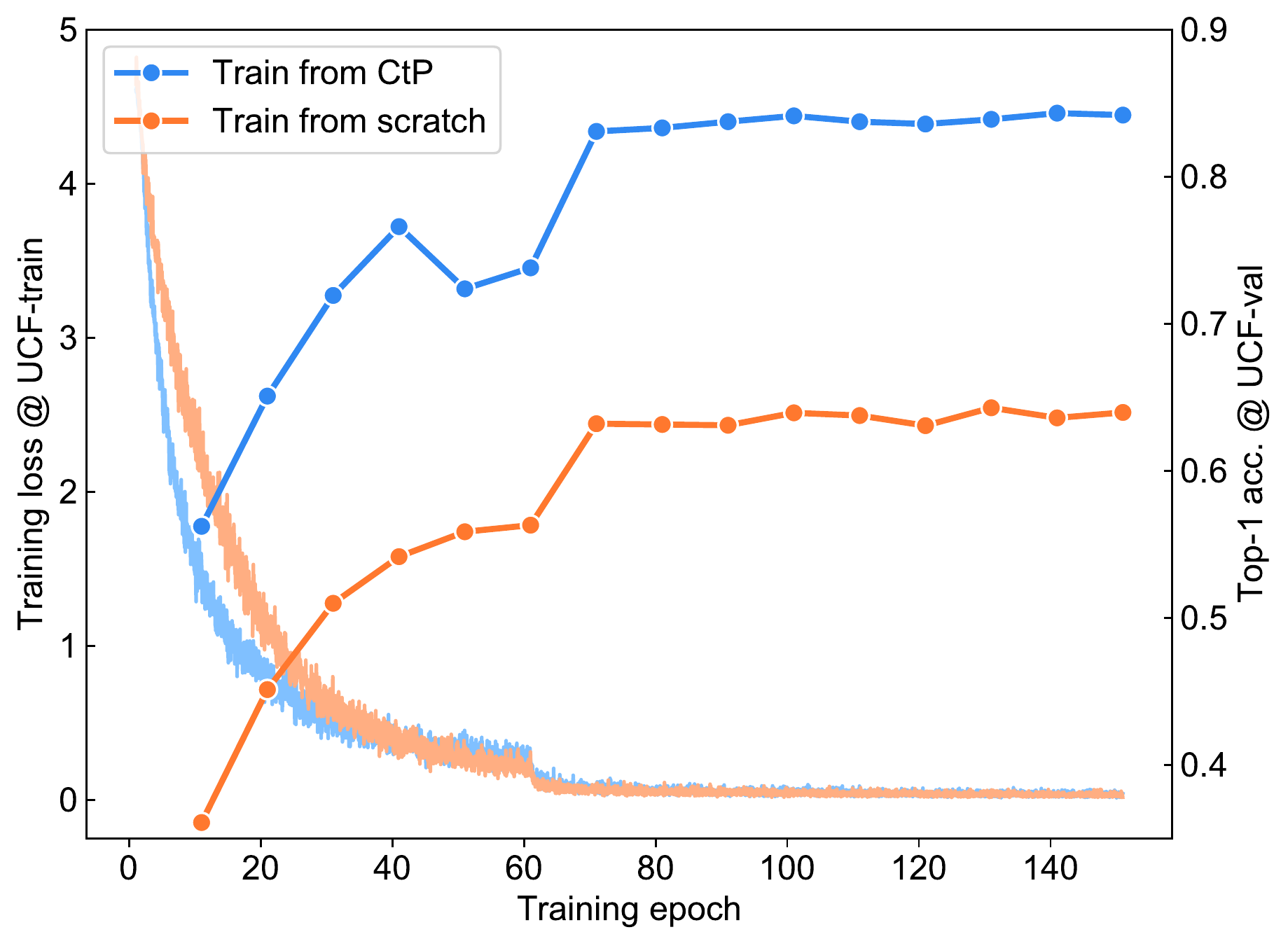}
\caption{Training loss curves and validation accuracy curves on UCF-101 dataset. The training process of action recognition task lasts for 150 epochs. We test the classifier in every 10 epochs.}
\label{FigLossCurve}
\end{figure}

\noindent\textbf{Video clip retrieval}\quad In the video clip retrieval task, the pretrained CNN encoder is directly used without further finetuning. We follow the same implementation as in VCOP \cite{VCOP}. The output feature of the CNN encoder is transferred to a fixed shape of $2\times3\times3\times512$ by an adaptive pooling operation, where each dimension denotes the temporal size, height, width and channels, respectively. For each video, we uniformly sample 10 clips and use the averaged feature of these 10 clips to represent the video-level feature. The cosine similarity between two different videos is considered as the distance metric. If a test video shares the same class label with one of its $k$-nearest training videos, it will be viewed as a correct retrieval.

\section{Analysis of Action Recognition}

In this section, we have a closer look at the training process of action recognition tasks. Generally, we consider two configurations when building a classifier on the UCF-101 dataset: (1) initialization from the CtP-learned model or (2) training from scratch. The training loss curve and test accuracy of both settings are shown in Figure \ref{FigLossCurve}. Compared with the baseline, our CtP pretraining enables a faster convergence speed in the early epochs. More importantly, even if both settings achieve close values of training losses in the final epoch, our CtP-pretrained model is significantly better than the baseline model in terms of validation accuracy. This is because our CtP game introduces motion priors about object movements, which helps the model quickly capture the moving targets and prevent overfitting.

\begin{table}[h]
\centering
\caption{Evaluation on the validation set of Kinetics-400.}\label{TableK400}
\begin{tabular}{@{}c|c|cc@{}}
\toprule
\multirow{2}{*}{Backbone} & \multirow{2}{*}{Pretraining} & \multicolumn{2}{c}{K400-val Acc. (\%)} \\
                          &                              & Top-1              & Top-5             \\ \midrule
R3D-18                       & None                         & 64.1               & 85.3              \\
R3D-18                       & CtP                          & \textbf{65.0}               & \textbf{85.7}              \\ \bottomrule
\end{tabular}
\end{table}

\begin{table}[h]
\centering
\caption{Ablation analysis on the backbone stride in the pretraining stage. The models are pretrained on the  UCF-101 dataset and transfer to the action recognition task.}\label{TableStride}
\begin{tabular}{@{}c|c|cc@{}}
\toprule
\multirow{2}{*}{Backbone} & \multirow{2}{*}{Stride} & \multicolumn{2}{c}{Top-1 Acc. (\%)} \\
                          &                              & UCF-101              & HMDB-51             \\ \midrule
R3D-18                       & 16                         & 83.5               & 52.5             \\
R3D-18                       & 8                          & \textbf{83.9}               & \textbf{53.6}              \\ \bottomrule
\end{tabular}
\end{table}

\begin{table}[h]
\centering
\caption{Ablation analysis on the temporal squeezing operation. The models are pretrained on the UCF-101 dataset and transfer to the action recognition task.}\label{TableSqueeze}
\begin{tabular}{@{}c|c|cc@{}}
\toprule
\multirow{2}{*}{Backbone} & \multirow{2}{*}{Squeeze?} & \multicolumn{2}{c}{Top-1 Acc. (\%)} \\
                          &                              & UCF-101              & HMDB-51             \\ \midrule
R3D-18                       &                          & 82.9               & 52.3 \\
R3D-18                       & \checkmark                  & \textbf{83.9}               & \textbf{53.6}              \\ \bottomrule
\end{tabular}
\end{table}

\section{Evaluation on Kinetics Dataset}


As suggested by the common practice, most of our experiments are conducted on UCF-101 and HMDB-51 datasets. However, these two datasets are relatively small (less than 10k videos). One may raise a natural concern, whether the CtP pretraining can still improve the performance when the downstream task has plenty of data. In order to address the concern, in this section, we transfer the CtP-pretrained model to the action recognition task on Kinetics-400 dataset. It has about 220k labeled videos for training and 18k videos for validation. 

The experimental results are presented in Table \ref{TableK400}. Our CtP-pretrained model outperforms the baseline model by a considerable margin. It proves that the downstream task on the large-scale dataset can still benefit from the CtP pretraining.

\section{Analysis of Backbone Stride}

In the pretraining stage, we introduce a slight modification to the backbone CNN encoder. The spatial stride of the last residual block is set to 1 instead of 2, i.e., the total spatial stride is decreased to 8. This strategy will lead to a finer feature map in the spatial dimensions, which has proven effective in object tracking literature. Note that the architecture is only modified in the pretraining stage. We reset the backbone is to the standard configuration for fair evaluations in the downstream tasks. The ablation results are presented in Table \ref{TableStride}. It clearly shows that this modification is helpful to video representation learning.

\section{Analysis of Scaling Factors}

When calculating the loss function, we apply four constant scaling factors $\mathbf{\sigma}$ to the prediction targets. It is a widely-used strategy in the area of object tracking and helps to balance the penalties of different regression terms. We provide the additional analysis on the choice of scaling factors, as shown in Table \ref{TableScaleFactor}. The performance is rather stable when the scaling factors are drawn from a reasonable range. 

\section{Analysis of Temporal Squeezing}

In our framework, we squeeze the temporal dimension of the extracted features before they are fed into the RoI align module. This operation can aggregate the information of the entire video and encourage the model to learn a more compact feature representation. As shown in Table \ref{TableSqueeze}, the action recognition results slightly degrade without this operation.

\begin{figure}[t!]
\centering
\includegraphics[width=1.0\linewidth]{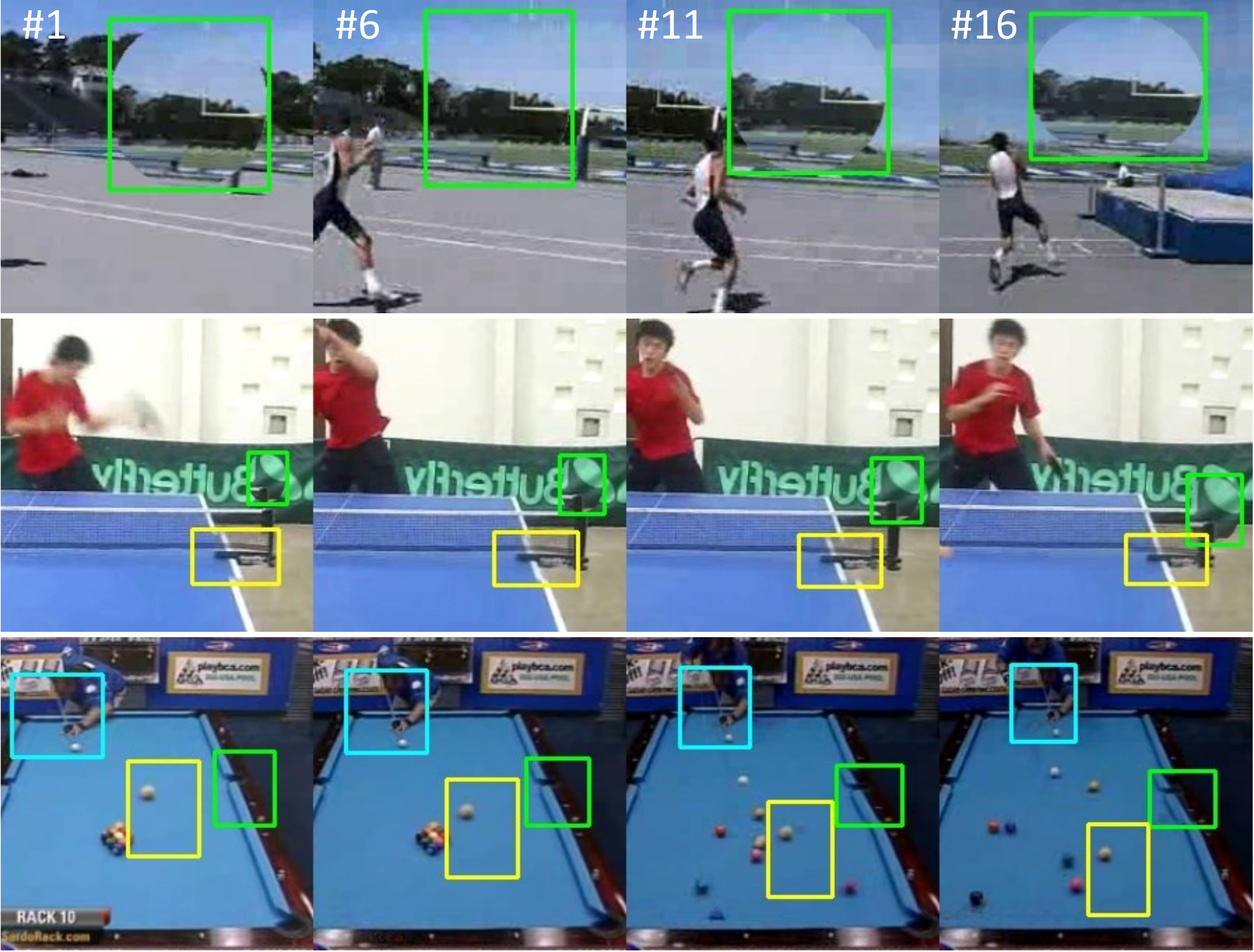}
\caption{Generated training examples.}
\label{FigTrainingExample}
\end{figure}

\begin{figure*}[!t]
\centering
\includegraphics[width=1.0\linewidth]{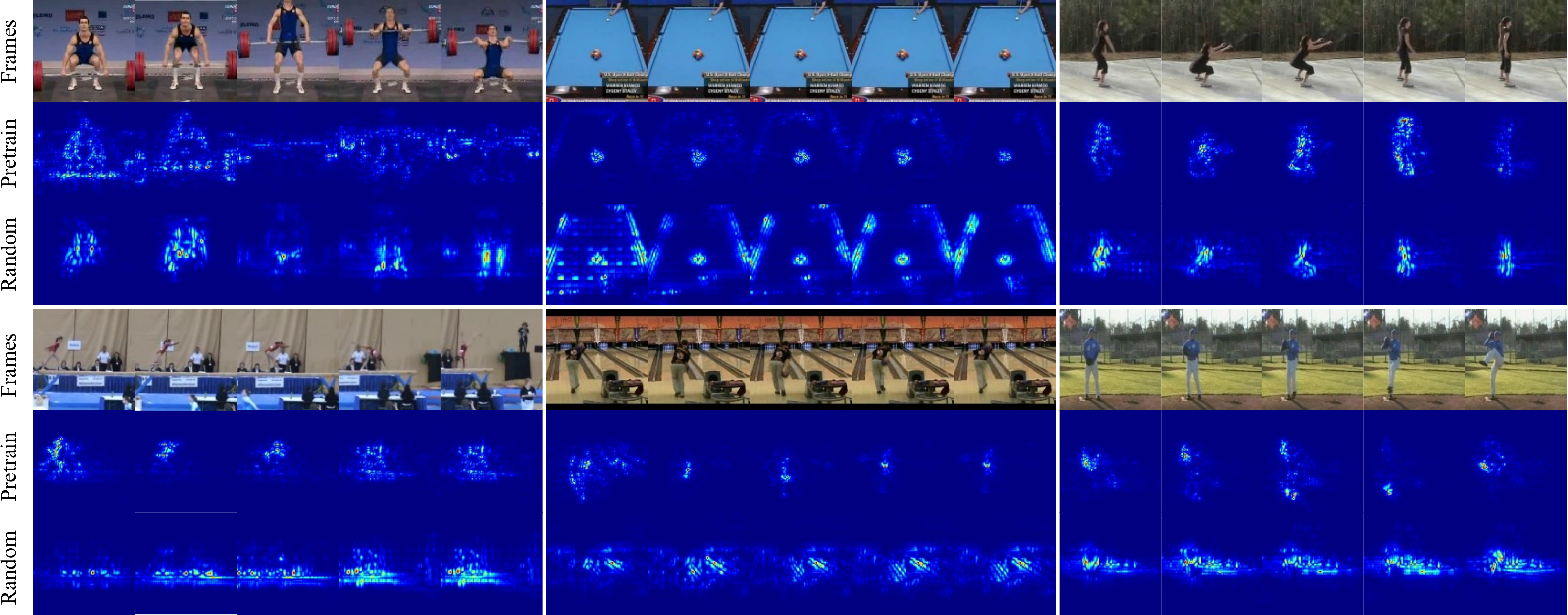}
\caption{More visualization examples. The important pixels are highlighted by Guided GradCAM algorithms \cite{GradCAM}.}
\label{FigMoreExamples}
\end{figure*}

\section{Details of Trajectory Synthesis}

In this section, we present the details of trajectory synthesis. Firstly, we determinate the bounding box location in the starting frame, which is randomly picked within a size range of $[16, 64]$ and an aspect ration range of $[1/2, 2]$. Secondly, we choose several key frames (3 to 5 in our experiments), and decide the bounding boxes in these key frames. To avoid the vast change of the synthetic trajectory, we add two constraints, namely speed constraint and scale constraint. For the former one, we force the position differences between the boxes in two neighbour key frames to be less than $3\Delta T$ pixels, where $\Delta T$ is the difference of two frame indexes. While for the scale constraint, the size ratio should lie in a range of $[\exp(-0.025\Delta T), \exp(0.025\Delta T)]$. Finally, the bounding boxes in the rest of frames are linearly interpolated between two key frames. Once we have the trajectory of bounding boxes, we can copy and paste the image patch to fill this trajectory. In Fig. \ref{FigTrainingExample}, we provide some generated examples, with one to three synthetic trajectories.

We also analyzes how the hyper-parameters of two constraints affect the pretraining performance. As shown in Table \ref{TableSpeedScale}, the performances are harmed if the speed is too fast or too slow. Besides, CtP pretraining is not sensitive to the scale changes.

\begin{table}[t]
\small
\setlength\tabcolsep{2.5pt}
\centering
\begin{tabular}{|c|l|c|c|c|c|}
\hline
\multicolumn{2}{|l|}{scaling factors} & \textcolor{blue}{(0.8/0.04)} & (0.8/0.08) & (0.8/0.02) & (1.0/0.05) \\ \hline
Cls. Acc.            & U101           & \textcolor{blue}{83.9}       & 83.3       & 83.6       & 83.2       \\ \cline{2-6} 
(\%)                 & H51            & \textcolor{blue}{53.6}       & 52.9       & 55.0       & 54.8       \\ \hline
\end{tabular}
\caption{Ablation analysis on the impact of scaling factors.} \label{TableScaleFactor} 
\end{table}
\begin{table}[t]
\small
\setlength\tabcolsep{2pt}
\centering
\begin{tabular}{|l|c|c|c|c|}
\hline
Speed & 1    & \textcolor{blue}{3}    & 5    & 15   \\ \hline
U101  & 80.6 & \textcolor{blue}{83.9} & 83.4 & 80.6 \\ \hline
H51   & 48.6 & \textcolor{blue}{53.6} & 54.0 & 49.3 \\ \hline
\end{tabular}
\begin{tabular}{|l|c|c|c|c|}
\hline
Scale & 0.010    & \textcolor{blue}{0.025}    & 0.040    & 0.100   \\ \hline
U101  & 83.3 & \textcolor{blue}{83.9} & 83.5 & 84.3 \\ \hline
H51   & 54.3 & \textcolor{blue}{53.6} & 53.2 & 54.2 \\ \hline
\end{tabular}
\caption{Ablation analysis on the patch speed and scaling ratio (described in L543.). The \textcolor{blue}{blue} columns denote the default setting.} 
\label{TableSpeedScale}
\end{table}

\section{More Visualizations}

Due to the space limitation, we only show two sequences in Section 5.7. Here, we further present more visualization examples in Figure \ref{FigMoreExamples}. The classifier trained from the CtP model can well capture the salient targets and avoid overfitting to the background regions.

{\small
\bibliographystyle{ieee_fullname}
\bibliography{draft_bib}

\begin{thebibliography}{10}\itemsep=-1pt

\bibitem{SpeedNet}
Sagie Benaim, Ariel Ephrat, Oran Lang, Inbar Mosseri, William~T Freeman,
  Michael Rubinstein, Michal Irani, and Tali Dekel.
\newblock Speednet: Learning the speediness in videos.
\newblock In {\em CVPR}, pages 9922--9931, 2020.

\bibitem{I3D}
Joao Carreira and Andrew Zisserman.
\newblock Quo vadis, action recognition? a new model and the kinetics dataset.
\newblock In {\em CVPR}, pages 6299--6308, 2017.

\bibitem{SimCLR}
Ting Chen, Simon Kornblith, Mohammad Norouzi, and Geoffrey Hinton.
\newblock A simple framework for contrastive learning of visual
  representations.
\newblock {\em arXiv preprint arXiv:2002.05709}, 2020.

\bibitem{PlaybackSpeed}
Hyeon Cho, Taehoon Kim, Hyung~Jin Chang, and Wonjun Hwang.
\newblock Self-supervised spatio-temporal representation learning using
  variable playback speed prediction.
\newblock {\em arXiv preprint arXiv:2003.02692}, 2020.

\bibitem{BERT}
Jacob Devlin, Ming-Wei Chang, Kenton Lee, and Kristina Toutanova.
\newblock Bert: Pre-training of deep bidirectional transformers for language
  understanding.
\newblock In {\em NAACL-HLT}, 2019.

\bibitem{VDIM}
R Devon~Hjelm and Philip Bachman.
\newblock Representation learning with video deep infomax.
\newblock {\em arXiv preprint arXiv:2007.13278}, 2020.

\bibitem{CutOut}
Terrance DeVries and Graham~W Taylor.
\newblock Improved regularization of convolutional neural networks with cutout.
\newblock {\em arXiv preprint arXiv:1708.04552}, 2017.

\bibitem{FlowNet}
Alexey Dosovitskiy, Philipp Fischer, Eddy Ilg, Philip Hausser, Caner Hazirbas,
  Vladimir Golkov, Patrick Van Der~Smagt, Daniel Cremers, and Thomas Brox.
\newblock Flownet: Learning optical flow with convolutional networks.
\newblock In {\em ICCV}, pages 2758--2766, 2015.

\bibitem{VKITTI}
Adrien Gaidon, Qiao Wang, Yohann Cabon, and Eleonora Vig.
\newblock Virtual worlds as proxy for multi-object tracking analysis.
\newblock In {\em Proceedings of the IEEE conference on computer vision and
  pattern recognition}, pages 4340--4349, 2016.

\bibitem{SomethingSomething}
Raghav Goyal, Samira~Ebrahimi Kahou, Vincent Michalski, Joanna Materzynska,
  Susanne Westphal, Heuna Kim, Valentin Haenel, Ingo Fruend, Peter Yianilos,
  Moritz Mueller-Freitag, et~al.
\newblock The" something something" video database for learning and evaluating
  visual common sense.
\newblock In {\em ICCV}, volume~1, page~5, 2017.

\bibitem{DPC}
Tengda Han, Weidi Xie, and Andrew Zisserman.
\newblock Video representation learning by dense predictive coding.
\newblock In {\em ICCV Workshop}, 2019.

\bibitem{MemDPC}
Tengda Han, Weidi Xie, and Andrew Zisserman.
\newblock Memory-augmented dense predictive coding for video representation
  learning.
\newblock {\em arXiv preprint arXiv:2008.01065}, 2020.

\bibitem{CoCLR}
Tengda Han, Weidi Xie, and Andrew Zisserman.
\newblock Self-supervised co-training for video representation learning.
\newblock {\em arXiv preprint arXiv:2010.09709}, 2020.

\bibitem{MoCo}
Kaiming He, Haoqi Fan, Yuxin Wu, Saining Xie, and Ross Girshick.
\newblock Momentum contrast for unsupervised visual representation learning.
\newblock In {\em CVPR}, pages 9729--9738, 2020.

\bibitem{MaskRCNN}
Kaiming {He}, Georgia {Gkioxari}, Piotr {Dollar}, and Ross {Girshick}.
\newblock Mask r-cnn.
\newblock {\em T-PAMI}, 42(2):386--397, 2020.

\bibitem{ResNet}
Kaiming He, Xiangyu Zhang, Shaoqing Ren, and Jian Sun.
\newblock Deep residual learning for image recognition.
\newblock In {\em CVPR}, pages 770--778, 2016.

\bibitem{KCF}
Joao~F. {Henriques}, Rui {Caseiro}, Pedro {Martins}, and Jorge {Batista}.
\newblock High-speed tracking with kernelized correlation filters.
\newblock {\em T-PAMI}, 37(3):583--596, 2015.

\bibitem{GOT10k}
Lianghua Huang, Xin Zhao, and Kaiqi Huang.
\newblock Got-10k: A large high-diversity benchmark for generic object tracking
  in the wild.
\newblock {\em TPAMI}, 2019.

\bibitem{TemporalTrans}
Simon Jenni, Givi Meishvili, and Paolo Favaro.
\newblock Video representation learning by recognizing temporal
  transformations.
\newblock {\em arXiv preprint arXiv:2007.10730}, 2020.

\bibitem{3DRotNet}
Longlong Jing, Xiaodong Yang, Jingen Liu, and Yingli Tian.
\newblock Self-supervised spatiotemporal feature learning via video rotation
  prediction.
\newblock {\em arXiv preprint arXiv:1811.11387}, 2018.

\bibitem{Kinetics}
Will Kay, Joao Carreira, Karen Simonyan, Brian Zhang, Chloe Hillier, Sudheendra
  Vijayanarasimhan, Fabio Viola, Tim Green, Trevor Back, Paul Natsev, et~al.
\newblock The kinetics human action video dataset.
\newblock {\em arXiv preprint arXiv:1705.06950}, 2017.

\bibitem{STPuzzle}
Dahun Kim, Donghyeon Cho, and In~So Kweon.
\newblock Self-supervised video representation learning with space-time cubic
  puzzles.
\newblock In {\em AAAI}, volume~33, pages 8545--8552, 2019.

\bibitem{HMDB}
Hildegard Kuehne, Hueihan Jhuang, Est{\'\i}baliz Garrote, Tomaso Poggio, and
  Thomas Serre.
\newblock Hmdb: a large video database for human motion recognition.
\newblock In {\em ICCV}, pages 2556--2563, 2011.

\bibitem{CorrFlow}
Zihang Lai and Weidi Xie.
\newblock Self-supervised learning for video correspondence flow.
\newblock {\em arXiv preprint arXiv:1905.00875}, 2019.

\bibitem{JointTaskCorr}
Xueting Li, Sifei Liu, Shalini De~Mello, Xiaolong Wang, Jan Kautz, and
  Ming-Hsuan Yang.
\newblock Joint-task self-supervised learning for temporal correspondence.
\newblock In {\em Advances in Neural Information Processing Systems}, pages
  318--328, 2019.

\bibitem{TSM}
Ji {Lin}, Chuang {Gan}, and Song {Han}.
\newblock Tsm: Temporal shift module for efficient video understanding.
\newblock In {\em ICCV}, pages 7082--7092, 2019.

\bibitem{VCP}
Dezhao Luo, Chang Liu, Yu Zhou, Dongbao Yang, Can Ma, Qixiang Ye, and Weiping
  Wang.
\newblock Video cloze procedure for self-supervised spatio-temporal learning.
\newblock In {\em AAAI}, 2020.

\bibitem{ACC}
Shuang Ma, Zhaoyang Zeng, Daniel McDuff, and Yale Song.
\newblock Learning audio-visual representations with active contrastive coding.
\newblock {\em arXiv preprint arXiv:2009.09805}, 2020.

\bibitem{CPC}
Aaron van~den Oord, Yazhe Li, and Oriol Vinyals.
\newblock Representation learning with contrastive predictive coding.
\newblock {\em arXiv preprint arXiv:1807.03748}, 2018.

\bibitem{FasterRCNN}
Shaoqing {Ren}, Kaiming {He}, Ross {Girshick}, and Jian {Sun}.
\newblock Faster r-cnn: Towards real-time object detection with region proposal
  networks.
\newblock {\em T-PAMI}, 39(6):1137--1149, 2017.

\bibitem{GTAV}
Stephan~R Richter, Vibhav Vineet, Stefan Roth, and Vladlen Koltun.
\newblock Playing for data: Ground truth from computer games.
\newblock In {\em European conference on computer vision}, pages 102--118.
  Springer, 2016.

\bibitem{GradCAM}
Ramprasaath~R Selvaraju, Michael Cogswell, Abhishek Das, Ramakrishna Vedantam,
  Devi Parikh, and Dhruv Batra.
\newblock Grad-cam: Visual explanations from deep networks via gradient-based
  localization.
\newblock In {\em ICCV}, pages 618--626, 2017.

\bibitem{LearnFromGAN}
Ashish Shrivastava, Tomas Pfister, Oncel Tuzel, Joshua Susskind, Wenda Wang,
  and Russell Webb.
\newblock Learning from simulated and unsupervised images through adversarial
  training.
\newblock In {\em Proceedings of the IEEE conference on computer vision and
  pattern recognition}, pages 2107--2116, 2017.

\bibitem{UCF101}
Khurram Soomro, Amir~Roshan Zamir, and Mubarak Shah.
\newblock Ucf101: A dataset of 101 human actions classes from videos in the
  wild.
\newblock {\em arXiv preprint arXiv:1212.0402}, 2012.

\bibitem{CBT}
Chen Sun, Fabien Baradel, Kevin Murphy, and Cordelia Schmid.
\newblock Contrastive bidirectional transformer for temporal representation
  learning.
\newblock {\em arXiv preprint arXiv:1906.05743}, 3(5), 2019.

\bibitem{PCL}
Li Tao, Xueting Wang, and Toshihiko Yamasaki.
\newblock Self-supervised video representation using pretext-contrastive
  learning.
\newblock {\em arXiv preprint arXiv:2010.15464}, 2020.

\bibitem{C3D}
Du {Tran}, Lubomir {Bourdev}, Rob {Fergus}, Lorenzo {Torresani}, and Manohar
  {Paluri}.
\newblock Learning spatiotemporal features with 3d convolutional networks.
\newblock In {\em ICCV}, pages 4489--4497, 2015.

\bibitem{R3D}
Du Tran, Heng Wang, Lorenzo Torresani, Jamie Ray, Yann LeCun, and Manohar
  Paluri.
\newblock A closer look at spatiotemporal convolutions for action recognition.
\newblock In {\em CVPR}, pages 6450--6459, 2018.

\bibitem{TrackingColor}
Carl Vondrick, Abhinav Shrivastava, Alireza Fathi, Sergio Guadarrama, and Kevin
  Murphy.
\newblock Tracking emerges by colorizing videos.
\newblock In {\em Proceedings of the European conference on computer vision
  (ECCV)}, pages 391--408, 2018.

\bibitem{PacePrediction}
Jiangliu Wang, Jianbo Jiao, and Yun-Hui Liu.
\newblock Self-supervised video representation learning by pace prediction.
\newblock {\em arXiv preprint arXiv:2008.05861}, 2020.

\bibitem{UDT}
Ning Wang, Yibing Song, Chao Ma, Wengang Zhou, Wei Liu, and Houqiang Li.
\newblock Unsupervised deep tracking.
\newblock In {\em CVPR}, pages 1308--1317, 2019.

\bibitem{wang2015unsupervised}
Xiaolong Wang and Abhinav Gupta.
\newblock Unsupervised learning of visual representations using videos.
\newblock In {\em Proceedings of the IEEE international conference on computer
  vision}, pages 2794--2802, 2015.

\bibitem{CycleCorrespondence}
Xiaolong Wang, Allan Jabri, and Alexei~A Efros.
\newblock Learning correspondence from the cycle-consistency of time.
\newblock In {\em CVPR}, pages 2566--2576, 2019.

\bibitem{VCOP}
Dejing Xu, Jun Xiao, Zhou Zhao, Jian Shao, Di Xie, and Yueting Zhuang.
\newblock Self-supervised spatiotemporal learning via video clip order
  prediction.
\newblock In {\em CVPR}, pages 10334--10343, 2019.

\bibitem{VTHCL}
Ceyuan Yang, Yinghao Xu, Bo Dai, and Bolei Zhou.
\newblock Video representation learning with visual tempo consistency.
\newblock {\em arXiv preprint arXiv:2006.15489}, 2020.

\bibitem{CEP}
Xinyu Yang, Majid Mirmehdi, and Tilo Burghardt.
\newblock Back to the future: Cycle encoding prediction for self-supervised
  contrastive video representation learning.
\newblock {\em arXiv preprint arXiv:2010.07217}, 2020.

\bibitem{PRP}
Yuan Yao, Chang Liu, Dezhao Luo, Yu Zhou, and Qixiang Ye.
\newblock Video playback rate perception for self-supervised spatio-temporal
  representation learning.
\newblock In {\em CVPR}, pages 6548--6557, 2020.

\bibitem{MMAction}
Dahua~Lin Yue~Zhao, Yuanjun~Xiong.
\newblock Mmaction.
\newblock \url{https://github.com/open-mmlab/mmaction}, 2019.

\bibitem{CutMix}
Sangdoo Yun, Dongyoon Han, Seong~Joon Oh, Sanghyuk Chun, Junsuk Choe, and
  Youngjoon Yoo.
\newblock Cutmix: Regularization strategy to train strong classifiers with
  localizable features.
\newblock In {\em Proceedings of the IEEE International Conference on Computer
  Vision}, pages 6023--6032, 2019.

\bibitem{Mixup}
Hongyi Zhang, Moustapha Cisse, Yann~N Dauphin, and David Lopez-Paz.
\newblock mixup: Beyond empirical risk minimization.
\newblock In {\em ICLR}, 2018.

\bibitem{ProbSTFunsion}
Yizhou Zhou, Xiaoyan Sun, Chong Luo, Zheng-Jun Zha, and Wenjun Zeng.
\newblock Spatiotemporal fusion in 3d cnns: A probabilistic view.
\newblock In {\em Proceedings of the IEEE/CVF Conference on Computer Vision and
  Pattern Recognition}, pages 9829--9838, 2020.

\bibitem{MiCT}
Yizhou Zhou, Xiaoyan Sun, Zheng-Jun Zha, and Wenjun Zeng.
\newblock Mict: Mixed 3d/2d convolutional tube for human action recognition.
\newblock In {\em CVPR}, pages 449--458, 2018.

\end{thebibliography}
}

\end{document}